\documentclass[runningheads]{llncs}
\usepackage[T1]{fontenc}
\usepackage{url}
\usepackage{cite}
\usepackage{amsmath,amssymb,amsfonts}
\usepackage{algorithmic}
\usepackage{graphicx}
\usepackage{textcomp}
\usepackage{xcolor}
\def\BibTeX{{\rm B\kern-.05em{\sc i\kern-.025em b}\kern-.08em
    T\kern-.1667em\lower.7ex\hbox{E}\kern-.125emX}}
\usepackage[utf8]{inputenc}
\usepackage{cleveref}
\usepackage{booktabs}
\usepackage{multirow}

\usepackage{mathtools}
\DeclarePairedDelimiter\abs{\lvert}{\rvert}%

\usepackage{csquotes}
\usepackage{siunitx}
\usepackage{colortbl}
\usepackage{adjustbox}

\usepackage[graphicx]{realboxes}
\usepackage{subfig}

\newcommand{\mic}[1]{\textcolor{blue}{}}

\newcommand{\eg}{e.\,g.\@ }
\newcommand{\ie}{i.\,e.\@ }

\newcommand{\pq}{$P \times Q$ }

\newcommand{\omicron}{\mathrm{o}}

\begin{document}
\title{PDPK: A Framework to Synthesise Process Data and Corresponding Procedural Knowledge for Manufacturing}
\titlerunning{Synthesising Process Data and Corresponding Procedural Knowledge}

\author{Richard Nordsieck\inst{1} \and
	André Schweizer\inst{1}, Michael Heider\inst{2} \and
	Jörg Hähner\inst{2}}
\authorrunning{R. Nordsieck et al.}

\institute{XITASO GmbH IT \& Software Solutions, Augsburg, Germany
	\email{\{richard.nordsieck, andre.schweizer\}@xitaso.com} \\
	\and
	Organic Computing Group, University of Augsburg, Augsburg, Germany\\
	\email{\{michael.heider, joerg.haehner\}@uni-a.de}}
\maketitle              % typeset the header of the contribution
\begin{abstract}
	Procedural knowledge describes how to accomplish tasks and mitigate problems.
	Such knowledge is commonly held by domain experts, \eg operators in manufacturing who adjust parameters to achieve quality targets.
	To the best of our knowledge, no real-world datasets containing process data and corresponding procedural knowledge are publicly available, possibly due to corporate apprehensions regarding the loss of knowledge advances.
	Therefore, we provide a framework to generate synthetic datasets that can be adapted to different domains.
	The design choices are inspired by two real-world datasets of procedural knowledge we have access to.
	Apart from containing representations of procedural knowledge in Resource Description Framework (RDF)--compliant knowledge graphs, the framework simulates parametrisation processes and provides consistent process data.
	We compare established embedding methods on the resulting knowledge graphs, detailing which out-of-the-box methods have the potential to represent procedural knowledge.
	This provides a baseline which can be used to increase the comparability of future work.
	Furthermore, we validate the overall characteristics of a synthesised dataset by comparing the results to those achievable on a real-world dataset.
	The framework and evaluation code, as well as the dataset  used in the evaluation, are available open source.

\end{abstract}

\section{Introduction}
\label{sec:intro}

While learning systems and machine learning enjoy a streak of successes in academic and industrial settings, their lack of the ability to reason and to generalise to unseen data remain hindrances to several use cases in practice.
Including background knowledge, \eg through knowledge infusion or neurosymbolic learning, could provide one possible solution to these problems \cite{Kursuncu.2020}.
Often, knowledge graphs are employed to represent background knowledge and are embedded into low dimensional vector representations \cite{Bordes.2013, Wang.2017} to facilitate their use in other downstream tasks and learning systems \cite{Portisch.2022}.
While factual and conceptual knowledge---knowledge of terminologies, classifications and generalisations \cite{Krathwohl.2002}---is often encountered in knowledge graphs \cite{Buchgeher.2021,Noy.2019,Miller.1990,Bollacker.2008} procedural and metacognitive knowledge---that describe knowledge of techniques and when to apply them as well as contextual knowledge \cite{Krathwohl.2002}---is seldom researched \cite{Nordsieck.2022b}.
Consequently, knowledge graph embedding methods are primarily developed and benchmarked on knowledge graphs containing factual and conceptual knowledge.
This leads to a situation where potential downstream systems are only capable of including factual or conceptual knowledge.
In practice, however, procedural and metacognitive knowledge play a significant part in representing reasoning and problem mitigation strategies, \eg in manufacturing \cite{Hoerner.2020}.
Here, incorporating tacit operator knowledge presents itself as one alternative to improve learning systems, \eg for predictive quality use cases, which suffer from scarce data due to manually optimized production processes.

In this domain, however, procedural knowledge and its representations in knowledge graphs exhibit several characteristics that differentiate it from factual knowledge, \eg ternary or chained binary relations, which introduce indirections, as well as asymmetry and literals, depending on the modelling detail \cite{Nordsieck.2022b,Nordsieck.2022c}.
Also, the overall size of the knowledge graphs is significantly smaller than those of established knowledge bases, \eg Freebase \cite{Bollacker.2008} or WordNet \cite{Miller.1990}.
Consequently, we surmise that studying established knowledge graph embedding methods on knowledge graphs containing procedural knowledge is an interesting research topic which, to the best of our knowledge, lacks public datasets.
To address this issue, we present a framework for generating synthetic datasets in manufacturing contexts which is
\begin{enumerate}
	\item highly configurable to suit a multitude of manufacturing scenarios,
	\item modular to ensure adaptability to unforeseen scenarios, and
	\item synthesises process data according to underlying rules for which representations in procedural knowledge graphs are generated.
\end{enumerate}
While a detailed simulation of the reality is beyond the scope of this framework datasets can be synthesised that have  characteristics similar to those exhibited by real-world procedural knowledge.
These datasets can be used to enable research into the interplay of procedural knowledge and process data, \eg evaluating predictive quality systems, approaches of knowledge-infused machine learning, data-based knowledge extraction as well as embeddings of procedural knowledge.

Apart from presenting the framework, we investigate its capability to produce datasets with characteristics that are exhibited by procedural knowledge in real-world manufacturing processes.
To this end, we compare a synthesised dataset to one gained in a fused-deposition-modelling (FDM) process---an additive manufacturing technique---on a link prediction task as well as on a metric designed to indicate downstream applicability of embeddings.

Based on related works and the state of the art (see \Cref{sec:related}), \Cref{sec:dataset-syn} describes the underlying assumptions and inner workings of the dataset generation founded on an understanding of production processes according to \cite{Nordsieck.2021}.
Building on this, we present our benchmark dataset on which we evaluate several established knowledge graph embedding methodologies (see \Cref{sec:benchmark}).
While future work is presented in \Cref{sec:future}, \Cref{sec:conc} concludes this paper.

\section{Related Work}
\label{sec:related}
The parametrisation process in manufacturing, and a data-based method to extract procedural knowledge from production data, is presented in \cite{Nordsieck.2021} and refined in \cite{Nordsieck.2022}.
We represent the procedural knowledge, which is underlying the reasoning of experts while mitigating quality defects in knowledge graphs.
We utilize these formalisations to build the synthesised production and parametrisation process underlying the dataset generation in \Cref{sec:benchmark-ds}.
Building on this, \cite{Nordsieck.2022b} presented the notion of modelling patterns, in the following called representations, for procedural knowledge graphs capable of representing different levels of detail of the represented knowledge and developed a sum-based embedding aggregation method for procedural knowledge.
We utilize these representations, specifically
representations for quantified conclusions ($\hat{\rho}$) that model ternary relations present in procedural knowledge as chained (ch) binary relations with quantifications either modelled as entities (e) or literals (l) $r_{\hat{\rho},\text{ch},\text{e}}$ and $r_{\hat{\rho},\text{ch},\text{l}}$ \cite{Nordsieck.2022c}.
As the resulting indirections of relations are hard to capture for embedding methods, they proposed the idea of retaining unquantified high-level relations $\eta$ which resulted in $r_{\hat{\rho},\text{ch},\text{e},\eta}$ and $r_{\hat{\rho},\text{ch},\text{l},\eta}$, respectively.
Furthermore, they evaluated different variations of the embedding method initially presented in \cite{Nordsieck.2022b} on a procedural knowledge specific metric.

Furthermore, we strive for a more fundamental understanding of the embedding methods intended for link prediction utilised in \cite{Nordsieck.2022b,Nordsieck.2022c} by benchmarking the resulting dataset in a link prediction setting.
Link prediction is the task of predicting either head $h$ or tail $t$ of a triple $(h,r,t)$ \cite{Wang.2017} and has been extensively used \cite{Sun.2019,Abboud.2020,Trouillon.2016,Kristiadi.2019} to benchmark embeddings on knowledge graphs that contain mostly factual or conceptual knowledge, such as Freebase \cite{Bollacker.2008} or WordNet \cite{Miller.1990}.
However, critics have mentioned the inherent bias of the frequently used datasets \cite{Mohamed.2020,Rossi.2021b,Tabacof.2019} and questioned the capability of embedding methods to capture knowledge graph semantics \cite{Jain.2021}.

Synthetic industrial datasets are prevalent for image data \cite{Nguyen.2022}, but relatively rare for process data or knowledge graphs.
Industrial process data is simulated by \cite{Hoag.2007} and \cite{Jeske.2005}.
Hoag and Thompson~\cite{Hoag.2007} present a framework for generating industrial size datasets using a XML specification and additional constraints.
Jeske et al.\@~\cite{Jeske.2005} describe a system that \enquote{generates data using statistical and rule-based algorithms and [$\ldots$] semantic graphs that [contain] interdependencies between attributes}.
Regarding knowledge graphs, Linjordet and Balog \cite{Linjordet.2020} discuss the use of templates for knowledge graph generation and associated problems due to leakage across data splits, which informed our decision for a random train-test split on the generated knowledge graphs.
We rely on the findings of Libes et al.\@ \cite{Libes.2017}, who explore challenges and suggest desirable features for the generation of synthetic data for manufacturing scenarios.

\section{PDPK: A Framework to Synthesise Datasets for Manufacturing}
\label{sec:dataset-syn}
Production processes manufacture a product or part that satisfies a set of target criteria, often pertaining to quality characteristics.
As such, we can assume that the product is observable at least once in the overall production process which provides the opportunity to assess the current quality by inspecting the respective quality characteristics $q \in Q$.
The chosen values for process parameters, $p \in P$, form a so called parametrisation that governs the production process, thereby influencing the resulting quality.
If quality defects occur for one or more quality characteristics or a production process is initially started, a parametrisation process is initiated to mitigate these defects and to arrive at a suitable parametrisation \cite{Nordsieck.2021}.
The parametrisation process consists of observing the quality characteristics of the previous iteration and iteratively adjusting process parameters until the target criterium, consisting of a subset of quality characteristics to optimize, $Q_{opt} \subseteq Q$ and a threshold $t \in [0,1]$, for a score function \begin{equation*}
	\Phi(i,Q_{opt}) = \frac{\sum_{q_j \in Q_{opt}}\frac{\omicron_{i,j}-\min(d_{q_j})}{\abs{d_{q_j}}}}{\abs{Q_{opt}}},
\end{equation*} with process iteration $i$ and quantified quality characteristic $\omicron$, is met.

Of the features suggested by Libes et al.\@ \cite{Libes.2017}, we implement the features relevant to our specific scenario.
In particular, we address the following seven features: (1) \textbf{data hiding} since our data is predominantly intended to be used for training and validating machine learning algorithms, we generate a test set that is separate and can be considered hidden from the training set.
The generated process data is \emph{filtered} during the creation of contiguous process iterations.
Also, PDPK can be viewed as a black box generator from the perspective of the algorithms that initiate it and consume its results; (2) \textbf{data quality} by modelling \emph{unreliable} physical systems with the option to introduce noise into the synthesised production process; (3) \textbf{data types} by focusing on generating data relating to product quality; (4) \textbf{repeatability} by allowing the seeding of the complete generation pipeline. Furthermore, by conforming to the FAIR principles a replicable versioning of the framework code is achieved; (5) \textbf{reproducibility} by providing the exact versions of libraries required, the framework ensures that the generated data is reproducible, regardless of the specific setup; (6) \textbf{model type} since PDPK can be viewed as a \emph{special-purpose model} for the synthetisation of artificial production processes and their parametrisation processes; (7) \textbf{integration, interoperability \& standards} by using widely used formats and standard, \eg *.csv for process data and *.ttl for Ressource Description Framework (RDF) representations of the rule base we allow their integration and interoperability with consumers.

\subsection{Dataset Generation}
\label{sec:datasetgen}
\begin{figure}[]
	\centering
	\includegraphics[width=\linewidth]{"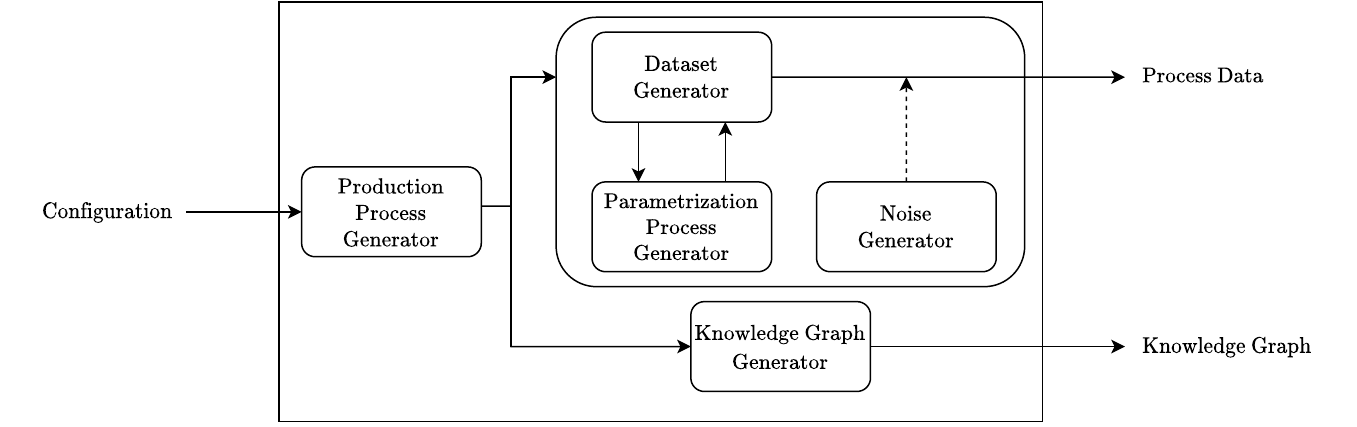"}
	\caption{High-level schema of PDPK.}
	\label{fig:schema}
\end{figure}

The framework for the generation of the dataset, PDPK, consists of multiple generators of synthesised production process, parametrisation process as well as components to facilitate the extraction of the underlying procedural knowledge and splitting the data into train and test sets as can be seen in \Cref{fig:schema}.

\subsubsection{Production Process}
To simulate the production process on which the para\-me\-tri\-sa\-tion process is conducted, we rely on separating the \pq space in disjunct parts.
The parameter $(P \times Q)_c$ defines which percentage of $(p,q)$-pairs should represent causal dependencies, \ie have functions that generate coherent values according to the selected configuration, \eg linear, quadratic or logarithmic.
Causal dependencies are further divided by the parameter $(P \times Q)_k$, which governs the percentage of causal combinations that are treated as known to the experts, \ie the underlying causal dependencies are available for exploitative behaviour during parametrisation processes and form the bulk of the knowledge graph.

Furthermore, the causal dependencies between $p$s and $q$s are constrained by the parameters $(p-q)_{\text{min}}$ and $(p-q)_{\text{max}}$, which denote the range from which the number of quality characteristics a given parameter affects is chosen.
Unknown causal dependencies govern the behaviour of the production process but have to be discovered by the simulated operator by exploration of the \pq space.
In any case, values for $p$s and $q$s are bound to ranges, $d$, representing their respective limits, just as real-world parameters are typically bound to some range.

As previously noted, each causal dependency is underpinned by a function $f_{k,j}: d_{p_k} \rightarrow d_{q_j}$.
For a process iteration $i$ the resulting quantified quality
\begin{equation*} o_{i,j} = \frac{\sum_{p_k\in P_{q_j}}f_{k,j}(\rho_{i,k})}{\abs{P_{q_j}}}
\end{equation*}
is calculated by averaging the influences of all quantified parameters $\rho$ for all $p \in P_{q_j}$ that affect $q_j$.

\subsubsection{Parametrisation Process}
A dataset is comprised of a set of multiple para\-me\-tri\-sa\-tion processes.
Each of these consists of a series of parametrisations and their resulting quality characteristics.
A randomly selected subset of quality characteristics is initially erroneous and consequently optimized over a non-deterministic number of iterations.
To mimic different degrees of operator knowledge, a certain percentage of the dataset's parametrisation processes consists of exploitative behaviour while the rest consists of explorative behaviour.
If the mitigation is attempted exploitatively, the process parameters are adjusted by subtracting the $\Delta$ calculated according to the following function from the previous value of $p$:
\begin{equation*}
	\Delta\rho_{i,k} = \frac{\sum_{q_j\in Q_{adj,p_k}}\Delta\rho_{i,k}^j}{\abs{Q_{adj,p_k}}},
\end{equation*} with $Q_{adj,p_k} = \{q_j | q_j \in Q_{opt} \land (p_k,q_j) \in (P \times Q)_k \}$ and \begin{equation*}
	\Delta\rho_{i,k}^j =
	\begin{cases}
		0,                                 & \text{if}\ \frac{\omicron_{i,j}-\min(d_{q_j})}{\min(d_{q_j})} \leq t \\
		(f_{k,j}^{-1})'(\omicron_{i-1,j}), & \text{otherwise}
	\end{cases}.
\end{equation*}

For parametrisation processes that are conducted according to the explorative behaviour, process parameters are adjusted independently by $\Delta\rho_{i,k}=-0.1\cdot \abs{d_{p_k}}\cdot \lambda$, with $\lambda$ chosen from $\{-1,1\}$ representing the direction, \ie increase, decrease, in which the process parameter is adjusted.
If the score improves, the previously adjusted parameter is further decreased until the threshold $t$ is reached.
If the score decreases, the parameter is adjusted in the opposite direction and if the score does not change, another process parameter is adjusted in the next iteration.

\subsubsection{Knowledge Graph Generation}
\label{sec:kggen}
\begin{figure}[]
	\centering
	\subfloat[Graphical illustration of representation $r_{\hat{\rho},\text{ch}, \text{e}, \eta}$. \label{fig:rep-qp-ch}]{%
		\includegraphics[width=0.45\textwidth]{"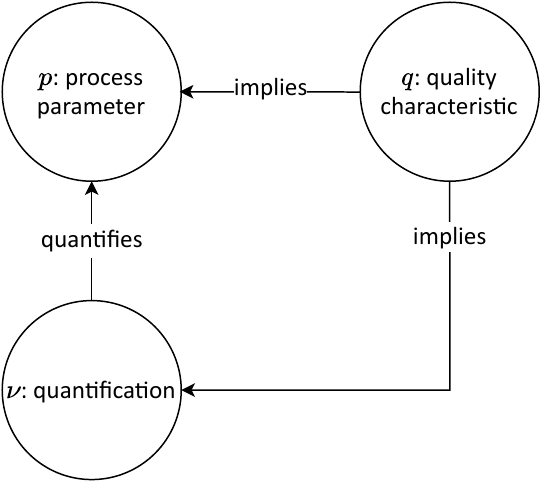"}%
	}\hfil
	\subfloat[KG for representation $r_{\hat{\rho},\text{ch}, \text{e}, \eta}$. \label{fig:rep-qp-rei}]{%
		\includegraphics[width=0.45\textwidth,
			height=5cm,
			keepaspectratio]{"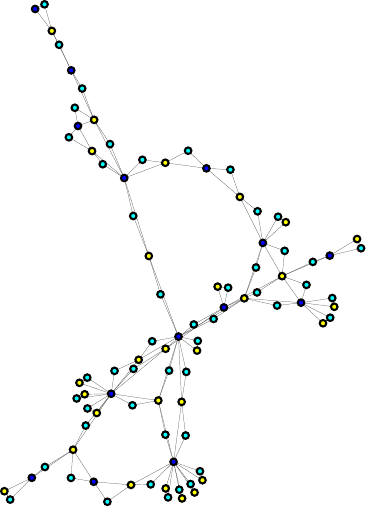"}%
	}
	\caption{Graphical illustration of representation $r_{\hat{\rho},\text{ch}, \text{e}, \eta}$ and the resulting knowledge graph.}
	\label{fig:graph-rep}
\end{figure}

Based on the underlying causal dependencies of $(p,q)$ pairs, $(P \times Q)_k$, knowledge graphs are generated depending on different representation patterns, presented in \cite{Nordsieck.2022b,Nordsieck.2022c}.
High-level $(p,q)$ relations (referred to as $\eta$) are straightforward since $q$ and $p$ form head and tail, respectively, which are connected with an \emph{implies} relation.
For quantified conclusions and conditions, \ie parameters and quality characteristics, the underlying functions have to be sampled to arrive at quantified parameters or quality characteristics, respectively.
Quantified parameters for a relation between $p_k$ and $q_j$ are determined by
\begin{equation}
	\label{eq:quantified-parameters}
	\hat{\rho}_{j,k}^{l,u} = \frac{\sum_{s=l}^{u-1}f_{k,j}^{-1}(s)-f_{k,j}^{-1}(s+1)}{\abs{d_{q_j}}-1},
\end{equation}
with $l=\min{(d_{q_j})}$ and $u=\max{(d_{q_j})}$, the lower and upper bounds of the range of the parameter, respectively.
Conditions, \ie quality characteristics, are quantified by determining ranges of $q$ for which different parameters should be applied, \eg by utilising estimators, such as \cite{Freedman.1981}, or providing a domain specific amount of ranges.
For each range, quantified parameters are calculated as described in \Cref{eq:quantified-parameters} with $l$ and $u$ the lower and upper limit of the respective range.
\Cref{fig:graph-rep} shows a graphical illustration of representation $r_{\hat{\rho},\text{ch}, \text{e}, \eta}$ and the corresponding KG of the benchmark dataset (see \Cref{sec:benchmark-ds}) gained by representing all rules using this representation.
In $r_{\hat{\rho},\text{ch}, \text{e}, \eta}$, the ternary relation is represented by a chained binary relation (see \Cref{fig:rep-qp-ch}).

\subsubsection{Train-Test Split}
\label{sec:testset}
Creating a train-test split on a dataset containing both, production data as well as accompanying knowledge graphs, comes with constraints to ensure consistency and include examples of knowledge in both train and test sets.
Methodically, one has the choice to split before actual process data has been generated, \ie on a causal dependency level, or to split after the fact.
Since the second approach is more closely related to established practices in the creation of test sets of link prediction \cite{Wang.2017} as well as downstream tasks, \ie predicting a variable based on the embedding, we follow the second approach.
To preclude train-test creep we explicitly prepare separate test sets for link prediction and process data settings.
The link prediction and downstream train and test sets are constituted by transferring a fixed percentage of relations from the graph or process iterations, respectively.
However, since generated knowledge graphs can contain process data, \eg for representations containing quantified process parameters and resulting quality characteristics, the corresponding knowledge graph has to be pruned of any process data that was part of iterations belonging to the train or test sets for the downstream train-test split.

\subsection{Usage}
The framework is a Python module that can be either used standalone or can easily be incorporated into existing data processing pipelines.
A multitude of parameters, \eg number of production process iteration, which directly influences the resulting size of the synthesised process data, and number of process parameters and quality characteristic, which indirectly influence the size of the generated knowledge graph, allow flexible configuration of the framework.
An overview of all parameters and their effect, as well as an introductory Jupyter notebook is given in the resource's user manual.

\subsection{Accordance to FAIR Principles}
The FAIR data principles \cite{Wilkinson.2016} describe a guideline to enhance the reusability of data resources with a focus on machine discover- and readability.
They govern findability, accessibility, interoperability and reusability.
In the following we discuss how PDPK conforms to these principles.
However, since the FAIR principles are not necessarily designed for frameworks for generating datasets and there is an ongoing discussion to their applicability to ontologies \cite{PovedaVillalon.2020} some principles are not directly applicable.

\paragraph{Findability:} PDPK is findable via the persistent Zenodo resource\footnote{\url{https://zenodo.org/record/7455849}} that archives the library from GitHub.
\paragraph{Accessibility:} As the focus of this resource is the framework and not specific datasets, endpoints to query the resulting KGs are not provided. However, these can be easily created using the generated RDF resources.
Since the KG represents only half of the generated dataset that contains both process data and procedural knowledge, it has no validity on its own.
\paragraph{Interoperability:} The framework utilises widely used formats, \ie *.json for configuration, *pickle, *.csv as well as *.ttl for produced artefacts and , *.rdf for meta data.
The vocabularies of existing ontologies in the manufacturing domain typically describe facts relating to products, resources or processes~\cite{Cao.2018} that are one abstraction level above those discussed in this paper.
Concerning processes, CDM-Core~\cite{Mazzola.2016} and ADACOR~\cite{Borgo.2007} contain concepts for machine \emph{setup} or \emph{launching} operations.
However, since parameter adjustments are also executed during production we do not differentiate between these two cases and consequently do not link to the vocabularies for these purposes.
Concerning products, MCCO~\cite{Usman.2011} contains the concept of rejected parts.
Related to that, CDM-Core introduces geometric flaws of different, fixed severities.
Quality characteristics as described in this paper, however, are not limited to geometric flaws and need to be expressed in a finer granularity.
Also, due to the abstraction level of the generated data, the concept of a product, the produced quantity or the quality of raw materials as present in ONTO-PDM~\cite{Panetto.2012} are not modelled in our case and are therefore not used.
However, depending on the specific application scenario different ontologies are likely to be prevalent which lowers the value of integrating our concepts with one of these ontologies in advance.

\paragraph{Reusability:} The framework is licensed via the very open MIT License and can therefore be adapted and reused in various contexts.

\subsection{Benchmark Dataset}
\label{sec:benchmark-ds}
In this section, we describe a dataset that we propose to serve as a benchmark for the performance of embedding methods on procedural knowledge and for downstream tasks utilizing the resulting embeddings.
We highlight which parameters were chosen based on the real-world datasets that form the basis for the synthetisation as well as characteristics of the resulting dataset.
Note, however, that it is not intended to serve as a direct approximation of one of these datasets but rather designed to reflect their more abstract characteristics.

\subsubsection{Parameters}
Using the parameters shown in \Cref{sec:datasetgen} and presented in the following, we are able to replicate characteristics of datasets from different manufacturing domains.
To arrive at benchmark data, however, we propose a parametrisation that is inspired by observations of manufacturing processes in both FDM as well as plastic extrusion, with the plastic extrusion process being significantly more complex than the FDM process, since multiple machines, \eg extruder, printer, laser, are coordinated in a production line.
In both cases, a significant fraction of possible $(p,q)$ combinations are not causally related since usually a $p$ or $q$ is only influenced by a small number of $q$s or $p$s, respectively.
E.g.\@ while there might be 20 different adjustable parameters, the surface finish quality characteristic is only dependent on material and air temperature but independent from the speed at which material is fed into the extruder.
Therefore, we set the number of $(p,q)$ combinations with causal dependencies, $(P \times Q)_c$, to 10\%.
Also, unknown influences, \eg of an environmental nature, are present in both scenarios.
To address this observation, we set $(P \times Q)_k=75\%$ to leave ample room for explorative behaviour.
The number of parameters, $\abs{P}$, is set to 46, while the number of qualities, $\abs{Q}$, is set to 16.
$(p-q)_{\min}$ and $(p-q)_{max}$ are set to 1 and 14, respectively.
Also the maximum length, \ie the number of process iterations, of a exploitative parametrisation process is set to 15 as in all but the most extreme cases operators are able to find a suitable parametrisation within 15 process iterations.
All modules' pseudo-random generators are seeded with 42.

\subsubsection{Properties}
The resulting dataset contains 500 process iterations, and 56 para\-metrisation processes, with the average parametrisation process requiring 8.93 ± 1.13 process iterations.
While the number of process iterations might appear small it is realistic in domains that produce a large variety of products at limited batch sizes.

\begin{table*}[h!]
	\centering
	\caption{Properties of the knowledge graphs resulting from the benchmark dataset for different representations.}
	\label{tab:dataset-properties}
	\begin{tabular}{lcccccccc}
		\toprule
		{Rep.}                                     & \#Edges & \#Vert. & \#Rel. & Close.\@ Cent. & Deg.\@ Cent. & Avg.\@ Neigh.\@ Deg. & Avg.\@ Deg. \\
		\midrule
		$r_{\hat{\rho},\text{ch},\text{e}}$        & 96      & 90      & 2      & 0.01±0.01      & 0.02±0.01    & 1.67±0.97            & 2.13±1.06   \\
		$r_{\hat{\rho},\text{ch}, \text{e}, \eta}$ & 144     & 90      & 2      & 0.02±0.02      & 0.04±0.03    & 4.11±2.41            & 3.2±2.47    \\
		$r_{\hat{\rho},\text{ch}, \text{l}, \eta}$ & 48      & 42      & 1      & 0.03±0.03      & 0.06±0.04    & 3.16±2.73            & 2.29±1.55   \\
		$r_{\hat{\rho},\text{rei}, \text{e}}$      & 144     & 138     & 3      & 0.01±0.01      & 0.02±0.01    & 1.83±0.47            & 2.09±1.2    \\
		\bottomrule
	\end{tabular}
\end{table*}

The properties of the resulting knowledge graph representations are presented in \Cref{tab:dataset-properties}.
Compared to other industrial knowledge graphs the number of edges, vertices and relations is relatively small \cite{Buchgeher.2021,Noy.2019}.
While this could be addressed by adjusting the dimensions of the \pq space accordingly, it is a characteristic of procedural knowledge that it is only available in limited quantities, especially compared to knowledge graphs containing information about products or production lines.
Also, the amount of edges and vertices decrease for representations that explicitly represent values as literals since these (and the relations to them) are counted separately.
In general, however, the more detailed a representation, the greater the amount of edges and vertices.
The average closeness centrality of each vertex is given by
\begin{equation*}
	C(x)=\frac{\abs{V}-1}{\sum_y d(y,x)},
\end{equation*}
with the number of vertices denoted by $\abs{V}$ and the distance $d$ between $x$ and $y$, is very small and further decreasing for the reification-based representations and the representation including conditions without high-level relations, $r_{\hat{\rho},\text{rei}, \text{e}}$ and $r_{\hat{\rho},\text{ch},\text{e}}$, respectively.
The average degree centrality for each vertex, \begin{equation*}
	C_D(x) = \frac{\abs{V}-1}{\text{deg}(x)},
\end{equation*}
with $\text{deg}(x)$ the degree, \ie the amount of in- and outgoing relations of $x$, is relatively low for all representations but further decreasing for representations $r_{\hat{\rho},\text{rei}, \text{e}}$ and $r_{\hat{\rho},\text{ch},\text{e}}$.
The average neighbour degree of each vertex, given by \begin{equation*}
	\frac{1}{\abs{N(i)}}\sum_{j\in N(i)} \text{deg}(j),
\end{equation*}
where $N(i)$ are the neighbours of vertex $i$, is at its highest for representations including the superfluous $\eta$ relation, which leads to a more interconnected graph with the degree of quality characteristics being higher which leads to the higher deviation.
The average degree shows a similar behaviour although less pronounced.

\subsubsection{Biases}

To analyse the knowledge graph of our dataset for sample selection biases \cite{Heckman.1979}, we evaluate it according to the biases presented in \cite{Rossi.2021b} (the description pertains to tail prediction, but is analogous for head prediction):
\begin{itemize}
	\item Type 1 Bias (B1): occurs if triples with relation r tend to always feature the same entity as tail, \ie a default answer for a relation exists.
	\item Type 2 Bias (B2): occurs if an entity is seen as head for a one-to-many or many-to-many relation. It tends to imply a constant entity as tail, \ie a default answer for a combination of relation and entity exists.
	\item Type 3 Bias (B3): occurs if there are two relations that tend to link the same entities, \ie if one could assume that one relation implies the other.
\end{itemize}
These biases are calculated for all triples, removing those where the bias is higher than the thresholds presented in \cite{Rossi.2021b}, \ie 0.5 for Type 1, 0.75 for Type 2 and 0.5 for Type 3.
We found that none of these biases are present in the dataset.

\section{Evaluation}
\label{sec:benchmark}
This section benchmarks established embedding methods on the synthetic benchmark dataset as well as the real-world FDM dataset.
This serves two purposes: Firstly, creating a baseline that can be used to ensure comparability of results of future work using PDPK; secondly showing PDPK's ability to accurately resemble characteristics of real-world datasets in regards to procedural knowledge.
We evaluate the embedding methods on a link prediction scenario and on aggregated sub-graphs.

\subsection{Experimental Setup}
To evaluate the link prediction performance, we rely on established evaluation metrics, \ie \emph{hits@k} and \emph{adjusted (arithmetic) mean rank index} (AMRI).
\emph{hits@k}, which describes the fraction of hits for which the entity appears under the first $k$ entries of the sorted list \cite{Berrendorf.2020}, is defined for the set of individual rank scores $\mathcal{I}$ as \begin{equation*}
	\text{hits@k} = \frac{\abs{\{r \in \mathcal{I} | r \leq k\}}}{\abs{\mathcal{I}}},
\end{equation*} with a value of 1 indicating an optimal score, whereas 0 is worst.
AMRI is better suited to compare different methods since it is not limited by $k$, taking the full sample into account.
It is defined as \begin{equation*}
	\text{AMRI} = \frac{2 \sum_{i=1}^n(r_i-1)}{\sum_{i=1}^n (\abs{\mathcal{S}_i})},
\end{equation*}
with $\mathcal{S}$ being the list of scores and $r$ the achieved rank.
Results of AMRI range in $[-1,1]$, where 0 indicates a performance similar to assigning random scores, 1 indicates optimal performance and values below 0 indicate worse than random performance \cite{Berrendorf.2020}.
The link prediction metrics are applied to the test set as described in \Cref{sec:testset}.

To evaluate whether the procedural knowledge contained in the graph is represented in the embeddings, we utilise a metric called \emph{matches@k}, which is \enquote*{based on the amount of overlap between the k closest quality characteristics in embedding and graph space} \cite{Nordsieck.2022b}.
In graph space, the closest quality characteristics are determined by ranking them according to their overlap in adjusted parameters, whereas in embedding space the euclidean distance is used on the associated sub-graphs.
These are achieved in a sum-based manner (see \cite{Nordsieck.2022b}) by propagating from the quality node for a representation specific number of steps.
Also, one has the option whether to include, $h$, the quality characteristic or not, $\bar{h}$.
We evaluate matches@k for $k=3$, which has been experimentally shown to be a good indicator for the FDM dataset.
Its results range in $[0,1]$, where $1$ indicates the best performance.
In the downstream scenario, generalisation of the learnt embeddings to previously unseen knowledge is not required as domain knowledge is unlikely to change.
Therefore, matches@k is evaluated in-sample.

For the link prediction setting, we investigate a set of well established knowledge graph embedding methods, namely TransE \cite{Bordes.2013}, ComplEx \cite{Trouillon.2016}, ComplEx-LiteralE, $\text{DistMult-LiteralE}_\text{g}$ \cite{Kristiadi.2019}, DistMult \cite{Yang.2014},  RotatE \cite{Sun.2019} and BoxE \cite{Abboud.2020}, which includes embedding methods of different complexity that could be able to capture indirections and other specific properties of the representations.
Additionally, we evaluate the downstream metric on RDF2Vec \cite{Ristoski.2016}.

We chose to train 46-dimensional embeddings, since this dimension allows its direct use in a downstream predictive scenario in our case.
The training was conducted using PyKEEN
\cite{Ali.2021b} with an Adam  optimizer with learning rate \num{4e-4} and weight decay \num{1e-5} for the link prediction methods and pyRDF2Vec
\cite{Vandewiele.2022} for RDF2Vec as well as the default parametrisations of the embedding methods.
The number of epochs---TransE: 400, ComplEx: 1000, ComplEx-LiteralE: 650, RotatE: 700, DistMult: 800, $\text{DistMult-LiteralE}_g$: 200, BoxE: 1500 and RDF2Vec: 1000---was chosen individually for each embedding method by inspecting its convergence on the train set.
30 runs were conducted to mitigate effects that could occur due to the random initialisation.

\begin{table*}[h!]
	\centering
	\caption{Link Prediction---Predicting Head Entities for benchmark and FDM datasets}
	\label{tab:results-lp-head}
	\begin{tabular}{llcccccc}
		\toprule
		                                                                         &         & \multicolumn{3}{c}{PDPK} & \multicolumn{3}{c}{FDM}                                                                               \\
		\rotatebox[origin=c]{0}{R.}                                              & Metric  & C.-LiteralE              & RotatE                  & BoxE                & C.-LiteralE         & RotatE    & BoxE                \\
		\midrule

		\multirow{4}{*}{\rotatebox[origin=c]{90}{$r_{\hat{\rho}, ch,e}$}}        & AMRI    & 0.91±0.02                & 0.64±0.11               & \textbf{0.94±0.01}  & 0.6±0.22            & 0.46±0.14 & \textbf{0.93±0.02}  \\
		                                                                         & Hits@1  & 0.16±0.07                & 0.16±0.07               & \textbf{0.24±0.06}  & 0.16±0.11           & 0.06±0.06 & \textbf{0.29±0.11}  \\
		                                                                         & Hits@5  & 0.53±0.09                & 0.36±0.09               & \textbf{0.65±0.08}  & 0.53±0.17           & 0.25±0.16 & \textbf{0.88±0.1 }  \\
		                                                                         & Hits@10 & 0.82±0.09                & 0.51±0.09               & \textbf{0.88±0.07}  & 0.68±0.2            & 0.43±0.14 & \textbf{0.99±0.03}  \\
		\arrayrulecolor{black!30}\midrule
		\multirow{4}{*}{\rotatebox[origin=c]{90}{$r_{\hat{\rho}, ch, e, \eta}$}} & AMRI    & 0.91±0.02                & 0.85±0.06               & \textbf{0.93±0.01}  & 0.8±0.12            & 0.79±0.08 & \textbf{ 0.93±0.03} \\
		                                                                         & Hits@1  & 0.13±0.05                & \textbf{0.21±0.07 }     & 0.18±0.06           & 0.26±0.13           & 0.31±0.09 & \textbf{ 0.33±0.12} \\
		                                                                         & Hits@5  & 0.48±0.09                & \textbf{0.61±0.08 }     & 0.6±0.07            & 0.7±0.17            & 0.76±0.08 & \textbf{ 0.89±0.06} \\
		                                                                         & Hits@10 & 0.79±0.08                & 0.78±0.08               & \textbf{0.87±0.05 } & 0.86±0.13           & 0.85±0.07 & \textbf{ 0.98±0.02} \\
		\arrayrulecolor{black!30}\midrule
		\multirow{4}{*}{\rotatebox[origin=c]{90}{$r_{\hat{\rho}, ch, l, \eta}$}} & AMRI    & \textbf{0.6±0.18 }       &                         &                     & \textbf{ 0.8±0.18 } &           &                     \\
		                                                                         & Hits@1  & \textbf{0.07±0.08}       &                         &                     & \textbf{ 0.41±0.18} &           &                     \\
		                                                                         & Hits@5  & \textbf{0.42±0.11}       &                         &                     & \textbf{ 0.81±0.18} &           &                     \\
		                                                                         & Hits@10 & \textbf{0.63±0.11}       &                         &                     & \textbf{ 0.87±0.14} &           &                     \\
		\arrayrulecolor{black!30}\midrule
		\multirow{4}{*}{\rotatebox[origin=c]{90}{$r_{\hat{\rho}, rei, e}$}}      & AMRI    & 0.91±0.02                & 0.87±0.03               & \textbf{0.94±0.01 } & 0.53±0.15           & 0.36±0.1  & \textbf{0.84±0.06}  \\
		                                                                         & Hits@1  & 0.29±0.08                & 0.83±0.03               & \textbf{0.84±0.03 } & 0.13±0.08           & 0.03±0.03 & \textbf{0.22±0.07}  \\
		                                                                         & Hits@5  & 0.72±0.07                & \textbf{0.84±0.03}      & \textbf{0.84±0.03 } & 0.43±0.11           & 0.15±0.07 & \textbf{0.76±0.09}  \\
		                                                                         & Hits@10 & 0.82±0.05                & \textbf{0.85±0.03}      & \textbf{0.85±0.03 } & 0.54±0.11           & 0.28±0.09 & \textbf{0.87±0.04}  \\
		\arrayrulecolor{black}\bottomrule
	\end{tabular}
\end{table*}

\begin{table*}[h!]
	\centering
	\caption{Subgraph Embedding Aggregation---Results, measured with matches@3 for benchmark and FDM datasets.}
	\label{tab:results-matches}
	\begin{tabular}{llcccccc}
		\toprule
		                                                                        &           & \multicolumn{3}{c}{PDPK} & \multicolumn{3}{c}{FDM}                                                                                     \\
		\rotatebox[origin=c]{0}{Rep.}                                           & H.        & C.-LiteralE              & BoxE                    & RDF2Vec            & C.-LiteralE        & BoxE               & RDF2Vec            \\
		\midrule

		\multirow{2}{*}{\rotatebox[origin=c]{0}{$r_{\hat{\rho}, ch,e}$}}        & $h$       & 0.24±0.06                & 0.34±0.05               & \textbf{0.42±0.05} & 0.44±0.07          & 0.51±0.05          & \textbf{0.66±0.04} \\
		                                                                        & $\bar{h}$ & 0.22±0.05                & 0.31±0.05               & \textbf{0.38±0.04} & 0.43±0.05          & 0.49±0.06          & \textbf{0.73±0.08} \\
		\arrayrulecolor{black!30}\midrule
		\multirow{2}{*}{\rotatebox[origin=c]{0}{$r_{\hat{\rho}, ch, e, \eta}$}} & $h$       & 0.3±0.06                 & 0.43±0.06               & \textbf{0.48±0.06} & 0.51±0.06          & \textbf{0.67±0.06} & 0.61±0.08          \\
		                                                                        & $\bar{h}$ & 0.27±0.06                & 0.38±0.05               & \textbf{0.44±0.05} & 0.55±0.06          & \textbf{0.63±0.06} & 0.58±0.08          \\
		\arrayrulecolor{black!30}\midrule
		\multirow{2}{*}{\rotatebox[origin=c]{0}{$r_{\hat{\rho}, ch, l, \eta}$}} & $h$       & \textbf{0.46±0.06 }      &                         &                    & \textbf{0.72±0.08} &                    &                    \\
		                                                                        & $\bar{h}$ & \textbf{0.42±0.07 }      &                         &                    & \textbf{0.58±0.07} &                    &                    \\
		\arrayrulecolor{black!30}\midrule
		\multirow{2}{*}{\rotatebox[origin=c]{0}{$r_{\hat{\rho}, rei, e}$}}      & $h$       & 0.21±0.05                & 0.24±0.05               & \textbf{0.38±0.05} & 0.4±0.1            & 0.43±0.07          & \textbf{0.67±0.07} \\
		                                                                        & $\bar{h}$ & 0.19±0.05                & 0.22±0.05               & \textbf{0.32±0.04} & 0.39±0.09          & 0.41±0.07          & \textbf{0.7±0.05}  \\
		\arrayrulecolor{black}\bottomrule
	\end{tabular}
\end{table*}

\subsection{Results}
We compare the results of evaluating different embedding methods for different representations of procedural knowledge on the benchmark dataset generated by PDPK and the real-world FDM dataset.
Our assumption is that while individual performance may vary, the overall characteristics are similar between the datasets.

\Cref{tab:results-lp-head} shows the results in a link prediction setting for head entities with the metrics AMRI and \emph{hits@k} with $k\in\{1,5,10\}$.
Explanations of the abbreviations used for representations can be found in \Cref{sec:related,sec:kggen}.
For representations including literals, results can only be reported for embedding methods which are able to embed literals.
The best result for each representation and metric is shown in bold.
For space reasons, only the best performing methods are shown.
In general, performance between the best embedding method of the two datasets is comparable, the exception being the literal based representation $r_{\hat{\rho}, ch, l, \eta}$.
The best performing method is the same in $87.5$\% of the cases, with BoxE a close second for hits@1 and hits@5 on $r_{\hat{\rho}, ch, e, \eta}$ of PDPK.

The results for the aggregated sub-graps as measured with matches@k are shown in \Cref{tab:results-matches}.
Here, the overall performance on the FDM dataset is consistently better with the best results per representation an average $27 \pm 4.6$\% above the benchmark dataset.
A plausible explanation for this behaviour is that the distribution of quality characteristics is narrower in the FDM case than the uniformly sampled PDPK, which directly translates to an increased matches@k result.
While the best embedding method per representation is more inconsistent than in the link prediction setting, they concur in $75$\% of cases.

Since the best method per representation concurs between benchmark and FDM datasets in at least $75\%$ of cases for link prediction and subgraph aggregation scenarios, we conclude that the dataset synthesised by PDPK is able to adequately represent characteristics of procedural knowledge.
The differences in overall performance as encountered in the sub-graph embedding evaluation are not detrimental to this conclusion, since the benchmark dataset does not aim to accurately replicate the FDM but rather show similar high-level characteristics.

\section{Future Work}
\label{sec:future}
An evaluation on a data-based knowledge extraction task and a downstream scenario such as knowledge infused learning \cite{Kursuncu.2020}, ideally on multiple datasets, is a logical next step to further validate the capability of PDPK to represent characteristics of real-world datasets.
The second would also help gaining a better understanding into the applicability of embedding methods originally intended for link prediction to downstream tasks for which \cite{Portisch.2022} provides theoretical considerations.
Also, investigating how knowledge graphs containing mixed representations and additional factual knowledge differ from knowledge graphs containing only one representation would provide a better indication for real-world tacit knowledge extracted with traditional, \eg interview-based, methods \cite{Hoerner.2022}.
A further interesting research topic is whether PDPK can be extended to simulate production processes to sufficient detail to allow the application of models trained on the synthetic dataset in real-world applications.
A first step in this direction would be modelling distributions for choosing the quality characteristics to optimise which would address the difference in results seen for matches@k.

In real world settings, knowledge is often associated with uncertainty.
As such, an extension of the dataset in this direction could increase its modelling capability.
However, in this case further requirements regarding the embedding methods are introduced, which necessitates the evaluation of specific embedding methods, \eg \cite{He.2015,Chen.2021,Vilnis.2018}.
\cite{Petzold.2021} compared the performance of text embeddings favourably with graph embeddings in an industrial context.
As such, it would be an interesting addition to the dataset to produce the expert knowledge not only in graph but also in textual form to evaluate its effect.
Furthermore, integration with existing industrial knowledge graphs or ontologies that contain detailed information pertaining to raw materials, the production process in general or the setup of the production line or machine could enable embedding methods and down-stream tasks to better grasp the specific application domain.

\section{Conclusion}
\label{sec:conc}
In this paper, we have presented PDPK, a framework for simulating parametrisation and underlying production processes in manufacturing.
This framework can be used to generate datasets consisting of knowledge graphs with procedural knowledge and process data.
These datasets can be used for link prediction and graph representation learning as well as downstream tasks, \eg predictive quality scenarios and knowledge extraction.
To the best of our knowledge, this is the first case of a publicly available dataset---and especially the first synthetic data generator---that combines process data with corresponding procedural knowledge.
This type of knowledge is highly relevant in many real-world applications and has different properties than factual and conceptual knowledge represented in datasets that are commonly used to evaluate link prediction methods.
We used PDPK to generate a benchmark dataset to compare several embeddings on a link prediction scenario, providing a reproducible baseline for future work using embedding methods on small scale procedural knowledge graphs.
We established, that BoxE and RDF2Vec achieved the most consistent results on link prediction and subgraph aggregation scenarios.
However, a \enquote{one-to-fit-all} approach is hard to justify, as has been previously noted \cite{Portisch.2022,Gogoglou.2020}.
Furthermore, we outlined possible extensions to PDPK and future research directions in which PDPK could be used.
We will continue to enhance PDPK during an ongoing research project and aim to build a community around using procedural knowledge in industrial contexts.
Due to the modular open source nature of PDPK it can be easily extended and adapted to different use cases.
As such, we hope PDPK is able to provide the basis for further research into representational learning with procedural knowledge and its combination with up- and downstream tasks.

\paragraph*{Resource Availability Statement:}
Source code for PDPK, the framework to generate datasets, as well as the benchmark dataset is available via Github\footnote{https://github.com/0x14d/PDPK} and the persistent url \url{http://purl.org/pdpk}.
This paper used PDPK 1.0.1 available via Zenodo\footnote{\url{https://zenodo.org/record/7919581}}.
Source code for reproducing the experiments is available via Github\footnote{\url{https://github.com/0x14d/embedding-operator-knowledge}}.
The FDM and plastic extrusion dataset cannot be made available as they are company confidential.

\bibliographystyle{IEEEtran}
\bibliography{citavirefs}

\end{document}